%% file: paper.tex
\documentclass{article}
\usepackage{spconf,amsmath,graphicx}
\usepackage{color}
\usepackage[pagebackref=false,breaklinks=true,letterpaper=true,colorlinks,bookmarks=false]{hyperref}
\usepackage{colortbl}
\usepackage{arydshln}
\usepackage{float}
\usepackage[table]{xcolor}

\newcommand{\equationref}[1]{Eq.~\ref{#1}}
\newcommand{\sectionref}[1]{Sec.~\ref{#1}}
\newcommand{\tableref}[1]{Tab.~\ref{#1}}
\newcommand{\figref}[1]{Fig.~\ref{#1}}

\newcommand{\eg}{\textit{e.g.}}
\newcommand{\ie}{\textit{i.e.}}
\newcommand{\argmax}{\operatornamewithlimits{argmax}}

\newcommand{\red}[1]{\textbf{\color[rgb]{1,0,0}{#1}}}
\newcommand{\blue}[1]{\textbf{\color[rgb]{0,0,1}{#1}}}

\title{INSTANCE SIGNIFICANCE GUIDED MULTIPLE INSTANCE BOOSTING FOR \\ ROBUST VISUAL TRACKING}
\name{Jinwu Liu$^{1,2,3}$ \quad Yao Lu$^{1,2}$\thanks{
        This work was supported in part by
        the National Natural Science Foundation of China (No.
        61273273), by the Research Fund for the Doctoral Program
        of Higher Education of China (No. 20121101110034) and by
        the Specialized Fund for Joint Building Program of Beijing
        Municipal Education Commission.
} \quad Tianfei Zhou$^{1,2}$}
\address{$^1$School of Computer Science, Beijing Institute of Technology \\
$^2$School of Information Science and Technology, Lingnan Normal University\\
$^3$Beijing Laboratory of Intelligent Information Technology\\
}

\begin{document}
%
\maketitle
\begin{abstract}
  \input{components/abstract}
\end{abstract}
\input{components/intro}
\input{components/method}
\input{components/experiments}

\input{components/conclusion}

\bibliographystyle{IEEEbib}
\bibliography{refs/refs}

\end{document}

%% file: components/abstract.tex
Multiple Instance Learning (MIL) recently provides an appealing way to alleviate the drifting problem in visual tracking.
Following the tracking-by-detection framework, an online MILBoost approach is developed
that sequentially chooses weak classifiers by maximizing the bag likelihood.
In this paper, we extend this idea towards incorporating the instance significance estimation into the online MILBoost framework.
First, instead of treating all instances equally,
with each instance we associate a significance-coefficient that represents its contribution to the bag likelihood.
The coefficients are estimated by a Bayesian formula that jointly considers the predictions from
multiple randomized MILBoost classifiers.
Next, we incorporate the estimates within a new boosting procedure for more effectively selection of weak classifiers.
Experiments with challenging public datasets show that
the proposed method outperforms both existing MIL based and boosting based trackers.

%% file: components/keywords.tex
Multiple Instance Learning, instance significance, online boosting, visual tracking

%% file: components/intro.tex
\section{Introduction}\label{sec:intro}

Tracking-by-detection has emerged as a leading approach for
accurate and robust visual tracking \cite{kalal2012tracking, babenko2011robust, zhang2013real, zhou2011online, avidan2007ensemble, grabner2006real, Kim_2015_ICCV,Hua_2015_ICCV,zhou2015locality,Khosla_2015_ICCV,zhou2015abrupt,li2017learning}.
This is primarily because it treats tracking as a detection problem, thereby avoiding modeling object dynamics
especially in the presence of abrupt motions \cite{zhou2014nearest} and occlusions \cite{pan2007robust}.
Tracking-by-detection typically involves training a classifier to detect the target in individual frames.
Once an initial detector is learned in the first frame,
the detector will progressively evolve to account for the appearance variations in both the target and its surroundings.

%
It is well known that accurate selection of training samples for the detector updating is rather significant for a successful tracking-by-detection method.
One common approach for this is to take the current tracking location as one positive example
and use the samples collected around the location for negatives.
While this simple approach works well in some cases,
the positive example used for detector updating may not be optimal if the tracking location is slightly inaccurate.
Over time this will degrade the performance of the tracker.
In contrast, many methods \cite{kalal2012tracking, avidan2007ensemble, grabner2006real} use multiple positive examples for updating,
where the examples are sampled from a small neighborhood around the current object location.

In principle, the latter updating scheme should be better because it exploits much more information.
However, as reported in existing literature,
it may confuse the appearance model since the label information about the positives is not precise.
Therefore, it may cause difficulties in finding an accurate decision boundary.
Consequently, a suitable algorithm needs to handle such sort of ambiguities in training data, especially in the positive ones.
Multiple Instance Learning (MIL) \cite{dietterich1997solving} can be exploited to achieve this goal,
since it allows for a weaker form of supervision to learn with instance label uncertainty.
For example, recent advances in object detection \cite{zhang2005multiple, ali2014confidence}
demonstrate that MIL is able to largely improve the detection performance.
Inspired by these applications,
Babenko et al. \cite{babenko2011robust} propose an online MILBoost approach to address the ambiguity problem in visual tracking.
Along with this thread, Zhang et al. \cite{zhang2013real} propose an online weighted MIL tracker,
and Bae et al. \cite{bae2014object} introduce structural appearance representation into the MIL based tracking framework.
In general, MIL enables these approaches to deal with slight appearance variations of the target during tracking,
in which case, most instances in the positive bag are relatively close to a true positive.
However, the trackers may fail in case of strong ambiguity, \eg, motion blur, pose change, etc.

To address this gap, in this work,
we follow the online boosting framework in \cite{grabner2006real} and propose a novel formulation of MILBoost for visual tracking.
The central idea behind our approach is learning the significance of instances, which we call \textit{significance-coefficients},
and incorporating them into the bag likelihood to guide the selection of weak classifiers in boosting.
%
In particular, we begin by building a group of randomized MILBoost learners,
and each provides its estimates for the instances being positive.
Assuming that the learners are independent, we show that the significance-coefficients can be easily estimated
through a Bayesian formulation.
Further, we introduce a variant of bag likelihood function based upon the significance-coefficients for the selection of weak classifiers.
%
%

%

%% file: components/method.tex
\section{Proposed Approach}\label{sec:method}

In the following, we first review the standard online multiple instance boosting method for tracking \cite{babenko2011robust}
and analyze its underlying limitations.
This analysis motivates then our new extension,
which allows for an accurate appearance model able to cope with diverse complex tracking scenarios.

\subsection{Online Multiple Instance Boosting}\label{sec:milboost}

Recently, Babenko et al. \cite{babenko2011robust} propose a novel online boosting algorithm for MIL (online MILBoost)
to address the example selection problem for adaptive appearance model updating in tracking.
In particular, given a training data set \( \{ (X_1, y_1), \ldots, (X_n, y_n) \} \) in current frame,
where a bag \( X_i = \{ x_{i1}, \ldots, x_{im} \} \) and \( y_i \in \{0, 1\} \) is its label,
as well as a pool of \(M\) candidate weak classifiers \( \mathcal{H} = \{h_1, h_2, \ldots, h_M\} \),
MILBoost sequentially chooses \(K\) weak classifiers from the candidate pool based upon the following criterion:
\begin{equation}\label{eq:boosting}
h_k = \argmax_{h \in \mathcal{H}} \mathcal{L}(\mathbf{H}_{k-1} + h)
\end{equation}
where \(\mathcal{L} = \sum_i (y_i \log p_i + (1-y_i)\log(1-p_i))\) is the log-likelihood over bags,
and \( \mathbf{H}_{k-1} = \sum_{i=1}^{k-1} h_i \) is the strong classifier consists of the first \( k-1\) weak classifiers.
Note that \(\mathcal{L}\) is the bag likelihood rather than instance likelihood used in traditional supervised learning approaches,
and \(p_i\) indicates the probability of bag \(i\) being positive, which is defined by the Noisy-OR model:
\begin{equation}\label{eq:noisy-or}
p_i = p(y_i | X_i) = 1 - \prod_j(1- p(y_i | x_{ij}))
\end{equation}
and \(p(y_i | x_{ij}) = \sigma(\mathbf{H}(x_{ij}))\) is the instance probability
where \(\sigma(x) = \frac{1}{1+e^{-x}} \) is the sigmoid function.

Note that the Noisy-OR model in \equationref{eq:noisy-or}, which is used to account for ambiguity, holds an assumption that
all instances in a bag contribute equally to the bag likelihood.
It is imprecise because according to the MIL formulation,
a positive bag contains at least one positive instance, but it may also contain many negative ones.
Clearly, the model in \equationref{eq:noisy-or} cannot identify the true positives in the positive bags.
While \cite{zhang2013real} mitigates this problem using a weighted MILBoost method,
we observe that slight inaccuracies in tracking results will lead to inaccurate weights,
thereby degrading the tracking performance.
Furthermore, not only is the likelihood model too restrictive,
but also one single MILBoost is not flexible enough for capturing the multi-modal distribution of the target appearance.

\subsection{Significance-Coefficients Estimation}

The previous analysis motivates our extension of standard MILBoost to a more robust model
so that it can handle various challenging situations.
Here we aim to integrate the instance significance into the learning procedure.
Note that our method is essentially different from \cite{zhang2013real}
because we in this work determine the instance significance discriminately
rather than simply weighting the instances according to Euclidean distances between the instances and the object location.


%
In particular, we begin with training \(N\) learners:
\begin{equation}
  \Phi = \{ \mathbf{H}_1, \ldots, \mathbf{H}_N \}
\end{equation}
where \(\mathbf{H}_i\) denotes a randomized MILBoost classifier learned in \sectionref{sec:milboost},
and the randomization is obtained by sampling different negative examples for each learner.
Then, for each instance \(x_{ij}\), its significance-coefficient \(r_{ij}\) is jointly determined by the predictions of the learners:
\begin{equation}
  r_{ij} = p(y_{ij} | \mathbf{H}_1, \ldots, \mathbf{H}_N)
\end{equation}
where \(y_{ij}\) denotes the label of \(x_{ij}\).
Assuming that the randomized MILBoost classifiers are conditional independent, we can rewrite the above formulation as:
\begin{align}
  r_{ij} & \propto p(\mathbf{H}_1, \ldots, \mathbf{H}_N | y_{ij})p(y_{ij}) \\
            & \propto p(y_{ij})\prod_{k=1}^N p(\mathbf{H}_k | y_{ij})
\end{align}
Note that we also have \( p(\mathbf{H}_k | y_{ij}) = \frac{p(y_{ij} | \mathbf{H}_k) p(\mathbf{H}_k)}{p(y_{ij})} \),
then the above formulation is equivalent to:
\begin{equation}\label{eq:r}
  r_{ij} \propto p(y_{ij})\prod_{k=1}^N \frac{p(y_{ij} | \mathbf{H}_k)}{p(y_{ij})}
\end{equation}
where \(p(y_{ij})\) is the prior indicating the probability that \(x_{ij}\) is positive, \ie, \(y_{ij} = 1\),
and \(p(y_{ij} | \mathbf{H}_k) = \sigma(\mathbf{H}_k(x_{ij})) \) is the prediction of \(\mathbf{H}_k\) over instance \(x_{ij}\).

\equationref{eq:r} has two characteristics in computing the significance-coefficients:
1) if the predicted probability \(p(y_{ij} | \mathbf{H}_k)\) is larger than the prior \(p(y_{ij})\),
the significance of \(x_{ij}\) will be enhanced;
2) considering the multiplicative part \(\prod_{k=1}^N p(y_{ij} | \mathbf{H}_k)\),
each predicted value can be viewed as imposing a weight to other predictions.
This intuitively benefits the significance estimation procedure.

Given the significance-coefficients of all instances in a positive bag,
we follow the underlying philosophy of MIL to estimate the bag significance:
\begin{equation}
  r_i = \max_j r_{ij}
\end{equation}
It should be noted that in MIL, ambiguity only exists in the positive bags.
Hence, we only estimate the significance-coefficients for instances in the positive bags,
but fix the significance of negative instances to \(r_{ij} = 1\), thus \(r_i = 1\).

\subsection{Refinement of Online MILBoost}
As introduced before, the Noisy-OR model is not precise
because it does not take the instance significance into account.
We now extend the Noisy-OR model in \equationref{eq:noisy-or} to the following:
\begin{equation}\label{eq:extend-noisy-or}
  p_i = p(y_i | X_i) = 1 - \prod_j(1 - p(y_i | x_{ij}))^{\alpha\frac{r_{ij}}{r_i}}
\end{equation}
The novel exponent term enables us to integrate the instance significance into \equationref{eq:noisy-or}.
In particular, the instance \(x_{ij}\) is equivalent to repeat \(\alpha \frac{r_{ij}}{r_i}\) times in the bags,
and \(\alpha\) is a constant that denotes the possible maximal repetition number for the instances.
In fact, in our experiments, we set \(\alpha = 1\) for the negative bags so that \equationref{eq:extend-noisy-or}
is equivalent to \equationref{eq:noisy-or}, and empirically set \(\alpha = 3 \) for the positive bags to incorporate instance significance.

Next, we develop an extended log-likelihood function over the bags as:
\begin{equation}
  \mathcal{L}_e = \sum_i r_i(y_i \log(p_i) + (1-y_i)\log(1-p_i))
\end{equation}
Given the new log-likelihood function,
we train a boosted classifier of weak learners as in \cite{babenko2011robust}:
%
\begin{equation}
  h_k = \argmax_{h \in \mathcal{H}} \mathcal{L}_e(\mathbf{H}_{e,k-1} + h)
\end{equation}
This is similar to the procedure in \equationref{eq:boosting},
except that we use a novel likelihood function \( \mathcal{L}_e \) instead of \(\mathcal{L}\) for weak classifier selection.
Finally, we obtain a strong classifier \(\mathbf{H}_e\) used as our discriminant appearance model.

\subsection{Weak Classifiers}

In this work, each object bounding box is represented using a set of Haar-like features \cite{viola2001rapid}.
Each feature consists of 2 to 4 rectangles, and each rectangle has a real-valued weight.
Thus, the feature value is a weighted sum of the pixels in all the rectangles.

For each Haar-like feature \(f_k\),
we associate it with a weak classifier \(h_k\) with four parameters \( (\mu_1, \sigma_1, \mu_0, \sigma_0) \):
\begin{equation}
  h_k(x) = \log \frac{p_t(y=1|f_k(x))}{p_t(y=0|f_k(x))}
\end{equation}
where \( p_t(f_t(x)| y=1) \sim \mathcal{N}(\mu_1, \sigma_1)  \) and similarly for \(y = 0\).
Note that the above equation establishes with a uniform prior assumption, \ie, \( p(y=1) = p(y=0) \).

Following \cite{babenko2011robust}, we update all the weak classifiers in parallel
when new examples \( \{ (x_1, y_1),\ldots, (x_n, y_n) \} \) are passed in:
\begin{align}
  \mu_1    &  \leftarrow \gamma \mu_1 + (1 - \gamma) \frac{1}{n} \sum_{i|y_i=1} f_k(x_i) \label{eq:u_mu} \\
  \sigma_1 &  \leftarrow \gamma \sigma_1 + (1 - \gamma) \sqrt{\frac{1}{n} \sum_{i|y_i=1} (f_k(x_i) - \mu_1)^2   }) \label{eq:u_sigma}
\end{align}
where \(\gamma \in [0, 1]\) is the learning rate.
The update rules for \(\mu_0\) and \(\sigma_0\) are similarly defined.
Note that our randomized MILBoost learners \(\Phi\) and the new classifier \(\mathbf{H}_e\) share the pool of candidate weak classifiers,
as well as the updating rules.

\subsection{Tracking Algorithm}

In this section, we summarize our tracking algorithm.
Without loss of generality, we assume the object location \(\ell_{t-1}^*\) at time \(t-1\) is given.
1) We first crop out some image patches \( X^{\gamma} = \{x: \parallel\ell(x) - \ell_{t-1}^*\parallel < \gamma \}\) as the positive instances,
and other ones \( X^\beta = \{x:  \gamma < \parallel\ell(x) - \ell_{t-1}^*\parallel < \beta \}\) as the negative instances,
where
\(\ell(x)\) denotes the location of patch \(x\) ,
\(\gamma\) and \(\beta\) are two scalar radius (measured in pixels).
2) Given the training examples, we learn a group of randomized MILBoost classifiers \( \Phi\)
as well as an improved MILBoost classifier \(\mathbf{H}_e\).
3) At time \(t\), we crop out a set of image patches \( X^s = \{x: \parallel\ell(x) - \ell_{t-1}^*\parallel < s \} \)
where \(s\) is a small search radius.
4) The object location \(\ell_t^*\) is ultimately obtained by:
\begin{equation}
  \ell_t^* = \ell\left(\argmax_{x\in X^s} p(y|x)\right)
\end{equation}
where \(p(y|x) = \sigma(\mathbf{H}_e(x))\) is the appearance model.
For other frames, our tracker repeats the above procedure to capture the object locations.

%% file: components/experiments.tex
\section{Experiments}\label{sec:experiments}

\input{figs/fig1}

\textbf{Datasets and Comparison Methods}
To evaluate the performance of the proposed algorithm thoroughly,
we perform experiments on nine public sequences with different challenging properties.
The total number of frames we tested is more than 9000.
We compare the method against 6 state-of-the-art algorithms, including 4 boosting-based (MIL \cite{babenko2011robust}, WMIL \cite{zhang2013real}, OAB \cite{grabner2006real},
and SBT \cite{grabner2008semi}) and 2 others (ASLA \cite{Lu2012Visual} and PCOM \cite{Wang2014Visual}).
For fair comparison, we run the source codes provided by the authors with default parameters.

\noindent
\textbf{Setup}
Our tracker is implemented in MATLAB and runs at 15 frames per second on a 2.93GHz Intel Core i7 CPU.
In the experiments,
the search radius \(s\) is set to 25 pixels, and
the scalars \(\gamma\) and \(\beta\) are set to \(4\) and \(50\) respectively.
For the negative image patches, we randomly select \( 200\) patches from \(X^\beta\).
Then, \(\Phi\) and \(\mathbf{H}_e\) are online updated using only 50 of 200 negative patches.
The number of randomized MILBoost classifiers is set to \( \parallel\Phi\parallel = 3 \),
and the learning rate in \equationref{eq:u_mu} and \equationref{eq:u_sigma} is fixed to \(\gamma = 0.85\).
Finally, the number of weak classifiers \(M\) is set to \(150\),
and each time \(K = 15\) classifiers are chosen to form a strong classifier.
\textit{Note that all the parameters are fixed for all sequences.}

\input{tables/cle_vor_merge}

\noindent
\textbf{Quantitative Evaluation:}
We employ \textit{Center Location Error (CLE)} and \textit{VOC Overlap Rate (VOR)} to evaluate the performance of our tracker.
CLE measures the position errors between central locations of the tracking results and the centers of the ground truth,
while VOR evaluates the overlap rate between the tracking results and the ground truth.
\tableref{table:merge} summarizes the average CLEs and VORs of all the seven trackers on the nine videos.
We can see that our tracker outperforms all boosting- and MIL-based trackers in these challenging sequences.
This is mainly because the multiple randomized classifiers enable us to capture the complex multi-modal distribution of the target appearance.
Besides, our bag likelihood function is more accurate than the previous algorithms.


\noindent
\textbf{Qualitative Evaluation:}
We evaluate the abilities of our method in dealing with various challenging factors,
and the results are illustrated in \figref{figure:result}.
(1) \textit{Motion Blur:} Sequences \textit{Boy} and \textit{Doll} contain fast motion and motion blur,
which result in large appearance variations.
The results in \figref{figure:result} demonstrate that incorporating instance significance into MILBoost
enables us to capture these variations.
(2) \textit{Low-Resolution Target:}
\textit{Panda} and \textit{Walking} show that our method copes well with the situations where the target is actually of low-resolution,
primarily because our method can select more discriminative features in the boosting stage than the previous approaches.
(3) \textit{Pose Change:}
The \textit{Sylv}, \textit{Girl} and \textit{Dog} sequences prove that our tracker is capable of handling pose changes and scale variations,
even though the scale of the bounding box is fixed in this work.
(4) \textit{Long-Term Tracking:} Finally, it's revealed in \textit{Dog} and \textit{Doll}
sequences that our tracker is more stable than other methods during long-term tracking,
owing that by incorporating significance-coefficients of instances,
our MIL method can well handle the ambiguity when updating the appearance model.

%% file: figs/fig1.tex
\begin{figure*}[t]
\begin{center}
  \includegraphics[width=.9\linewidth]{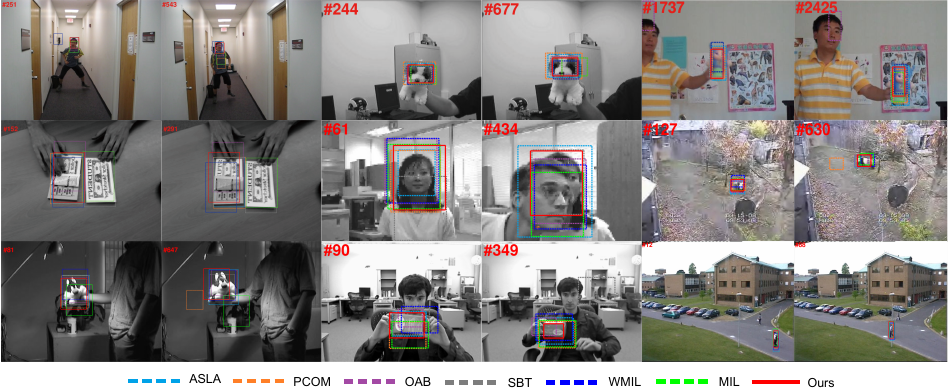}
\end{center}
\caption{Sample tracking results of the evaluated algorithms on nine challenging image sequences. The figures are placed in the same order as \tableref{table:merge}.}
\label{figure:result}
\end{figure*}

%% file: tables/cle_vor_merge.tex
\begin{table*}[!t]
    \caption{Tracking Results in terms of CLE and VOR. Top two results are shown in \red{red} and \blue{blue} fonts.}
    \tiny
\begin{center}
\resizebox{0.9\textwidth}{!}{%
  \rowcolors{2}{gray!25}{white}
  \begin{tabular}{|l|c:c|c:c|c:c|c:c|c:c|c:c|c:c|c:c|}
  \hline
  {} & \multicolumn{2}{c|}{\textbf{ASLA}} & \multicolumn{2}{c|}{\textbf{PCOM}} & \multicolumn{2}{c|}{\textbf{OAB}} & \multicolumn{2}{c|}{\textbf{SBT}}& 
  \multicolumn{2}{c|}{\textbf{WMIL}}& \multicolumn{2}{c|}{\textbf{MIL}} & \multicolumn{2}{c|}{\textbf{Ours}} \\
  {}               & CLE  & VOR  & CLE  & VOR  & CLE  & VOR  & CLE  & VOR  & CLE  & VOR & CLE  & VOR & CLE  & VOR \\
  \hline\hline
  Boy              & \blue{6.5}  & \blue{0.71} & 10.5 & 0.73 & 6.8  & 0.67 & 7.3  & 0.53 & 58.1 & 0.43 & 30.3 & 0.48 & \red{5.4}  & \red{0.78} \\
  Dog              & \red{6.4}  & \red{0.77} & 10.9 & 0.72 & 19.9 & 0.40 & 27.8 & 0.39 & 14.6 & 0.45 & 25.5 & 0.47 & \blue{10.2} & \blue{0.73} \\
  Doll             & \red{4.7}  & \red{0.78} & 9.0  & 0.70 & 19.6 & 0.59 & 12.1 & 0.64 & 41.5 & 0.39 & 30.2 & 0.34 & \blue{6.3}  & \blue{0.75} \\
  Dollar           & 9.1  & \blue{0.75} & \blue{6.4}  & \red{0.81} & 38.2 & 0.61 & 77.6 & 0.35 & 35.1 & 0.63 & 18.5 & 0.68 & \red{6.3}  & \red{0.81} \\
  Girl             & \red{15.6} & \red{0.74} & 46.6 & 0.32 & 25.0 & 0.54 & 18.0 & 0.71 & 54.4 & 0.44 & 38.8 & 0.41 & \blue{17.3} & \blue{0.72} \\
  Panda            & 8.8  & 0.65 & 45.0 & 0.11 & 8.2  & 0.67 & 7.2  & 0.70 & \red{6.3}  & \blue{0.71} & 7.8  & 0.76 & \blue{6.7}  & \red{0.73} \\
  Sylv             & \blue{9.8}  & \blue{0.62} & 61.5 & 0.19 & 18.7 & 0.65 & 17.0 & 0.64 & 19.9 & 0.60 & 44.7 & 0.43 & \red{6.5}  & \red{0.70} \\
  Twinings         & \blue{7.5}  & \blue{0.73} & 11.5 & 0.54 & 33.9 & 0.54 & 19.7 & 0.61 & 21.7 & 0.55 & 20.5 & 0.57 & \red{6.3}  & \red{0.81} \\
  Walking          & 5.5  & 0.70 & \blue{5.1}  & \blue{0.72} & 5.2  & 0.71 & 5.3  & 0.70 & 11.9 & 0.51 & 6.6  & 0.64 & \red{5.0}  & \red{0.74} \\
  \hline\hline
  \textbf{Average} & \blue{8.2} & \blue{0.71} & 22.9 & 0.54  & 19.5 & 0.60 & 21.3 & 0.59  & 29.3 & 0.52 & 22.5 & 0.53 & \red{7.8} & \red{0.75}    \\
  \hline
  \end{tabular}}
  \end{center}
  \label{table:merge}
\end{table*}

%% file: components/conclusion.tex
\section{Conclusion}\label{sec:conclusion}

Inspired from the recent success of multiple instance learning (MIL) in tracking,
we proposed a novel algorithm that incorporates the significance-coefficients of instances into 
the online MILBoost framework.
Our approach consists of two steps:
(i) significance-coefficients estimation via a Bayesian formulation
    based on the predictions given by the randomized MILBoost classifiers, and
(ii) a flexible scheme for incorporating the instance significance into the objective function of online MILBoost.
In the experiments, we evaluate our method on several publicly available datasets and the results show its better performance.